\documentclass[runningheads]{llncs}
\usepackage{graphicx}
\usepackage{amsmath}
\usepackage{amssymb}

\usepackage{array}
\newcolumntype{L}[1]{>{\raggedright\let\newline\\\arraybackslash\hspace{0pt}}m{#1}}
\newcolumntype{C}[1]{>{\centering\let\newline\\\arraybackslash\hspace{0pt}}m{#1}}
\newcolumntype{R}[1]{>{\raggedleft\let\newline\\\arraybackslash\hspace{0pt}}m{#1}}
\begin{document}
\title{Defending against adversarial attacks on medical imaging AI system, classification or detection?}
%
\author{Xin Li \and Deng Pan \and Dongxiao Zhu}
%
%
\institute{Department of Computer Science, Wayne State University, USA \\ \email{\{xinlee,pan.deng,dzhu\}@wayne.edu}}
%
\maketitle              
\begin{abstract}
Medical imaging AI systems such as disease classification and segmentation are increasingly inspired and transformed from computer vision based AI systems. Although an array of adversarial training  and/or loss function based defense techniques have been developed and proved to be effective in computer vision, defending against adversarial attacks on medical images remains largely an uncharted territory due to the following unique challenges: 1) label scarcity in medical images significantly limits adversarial generalizability of the AI system; 2) vastly similar and dominant fore- and background in medical images make it hard samples for learning the discriminating features between different disease classes; and 3) crafted adversarial noises added to the entire medical image as opposed to the focused organ target can make clean and adversarial examples more discriminate than that between different disease classes. In this paper, we propose a novel robust medical imaging AI framework based on Semi-Supervised Adversarial Training (SSAT) and Unsupervised Adversarial Detection (UAD), followed by designing a new measure for assessing systems adversarial risk. We systematically demonstrate the advantages of our robust medical imaging AI system over the existing adversarial defense techniques under diverse real-world settings of adversarial attacks using a benchmark OCT imaging data set.                 

\end{abstract}
\section{Introduction}
Deep neural networks (DNNs) have achieved significant advancement in various tasks of medical imaging, including pneumonia detection (X-ray) \cite{rajpurkar2017chexnet}, early diagnosis of prostate cancer (MRI) \cite{reda2018new}, retina diseases classification (OCT) \cite{eladawi2018classification} and nodule segmentation (CT) \cite{qin2019pulmonary}. To deploy DNN-based AI systems to support disease diagnosis in those clinical applications, the robustness of DNN models increasingly arises as a great importance. Recent studies \cite{finlayson2018adversarial,taghanaki2018vulnerability,paschali2018generalizability,ozbulak2019impact} have specifically explored the reliability of DNN models in both classification and segmentation tasks of medical imaging. They show that DNNs can suffer from significant performance drop when predicting adversarial samples \cite{szegedy2013intriguing}, which are intentionally crafted inputs with human imperceptible perturbations that can completely fool the trained DNN models. To generate adversarial samples, various types of attacks have been proposed, such as Fast Gradient Sign Method (FGSM) \cite{goodfellow2014explaining} and its variant with stronger attacks Projected Gradient Descent (PGD) \cite{madry2017towards}, and optimization-based attack Carlini \& Wagner (C\&W) \cite{carlini2017towards}. 
For medical imaging segmentation tasks, Ozbulak et al. \cite{ozbulak2019impact} propose an adaptive segmentation mask attack (ASMA), which produces a crafted mask to fool the trained DNN model. Such vulnerability of DNNs to adversarial samples has raised substantial safety concerns on the deployment of medical imaging AI systems at scale.

\begin{figure}[ht]
\centering
\includegraphics[height=4.5cm]{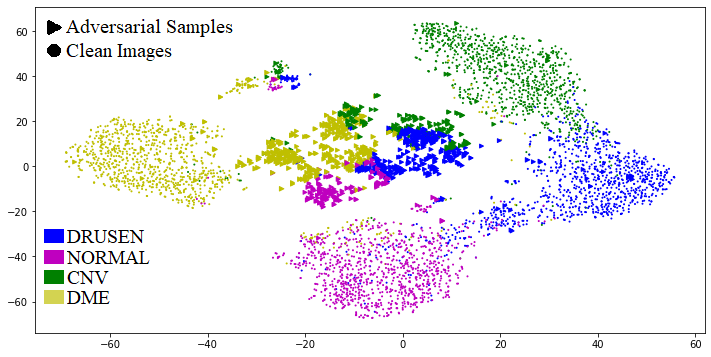}
\caption{T-SNE visualization of penultimate layer activations of the model trained on the OCT dataset \cite{kermany2018identifying}. The clean images are represented by solid circles with each color represents a true class. The adversarial samples (triangles) are crafted by PGD with a perturbation budget $\epsilon = 0.005$ where each color represents a predicted class. For each class, UAD is capable of filtering out the OOD adversarial samples (center) and SSAT enables the model to correctly predict in-distribution adversarial samples.}
\label{fig:vis}
\end{figure}

To defend against these adversarial attacks, different strategies have been developed. One major line of those methods is based on adversarial training (AT)\cite{goodfellow2014explaining,tramer2017ensemble}, which improves model's adversarial robustness by augmenting the training set with adversarial samples. However, AT for DNNs in medical imaging is problematic as they are primarily designed for natural images and requires a large labeled training set \cite{stanforth2019labels} whereas medical data sets are usually with a small amount of labeled samples. Another line of efficient defense approaches is to learn discriminative features for classifying natural and adversarial samples \cite{chen2019improving,taghanaki2019kernelized}. With large inter-class separability and intra-class compactness in latent feature space, attacks with a small perturbation budget are more difficult to succeed. However, medical imaging AI systems can be more susceptible to even benign attacks \cite{ma2019understanding} since medical images are highly standardized with well-established exposure and quality control, featuring a significant overlap in fore- and backgrounds. As a result, a small adversarial perturbation on the entire clean images can significantly distort their distribution in the latent feature space, which can be detrimental to the model performance on clean images. As shown in Figure \ref{fig:vis}, adversarial samples deviate significantly from the distribution of clean samples, implying that they are out-of-distribution (OOD) hard samples for supervised classification. Consequently the accurate prediction is not attainable yet unsupervised detection remains as a more promising path \cite{xin2020detection}.

Recently several techniques are proposed to improve the effectiveness of defensive methods for medical images. In segmentation tasks, He et al. \cite{he2019non} found that global contexts and global spatial dependencies are effective against adversarial samples, thus they propose a non-local context encoder in the medical image segmentation system to improve adversarial robustness. In classification tasks, Taghanaki et al. \cite{taghanaki2019kernelized} use a radial basis mapping kernel to transform and separate features on a manifold to diminish the influence of adversarial perturbation. Based on features extracted from a trained DNN model, Ma et al. \cite{ma2019understanding} attempt to distinguish adversarial samples from clean ones via density estimation in the subspace of deep features learned by a classification model. Although it achieves impressive performance, the so-called `detection' methods rely on estimating the density of adversarial samples, e.g., via local \cite{ma2018characterizing} or Bayesian uncertainty \cite{feinman2017detecting} approaches, consequently the effectiveness is limited to the attack methods that are previously seen.  


In order to train the AI system with a small set of labeled images to improve adversarial robustness against unseen and heterogeneous attacks, instead of performing supervised adversarial training, we take a different perspective via unsupervised detection of adversarial samples without the need for estimating density of the adversarial samples. We present a hybrid approach that enhances DNN defensive power using semi-supervised adversarial training (SSAT) and unsupervised adversarial detection (UAD). Specifically, we utilize both labeled and unlabeled data to generate psudo-labels for SSAT to improve the robustness of class prediction. To mitigate the distribution distortion of unseen adversarial samples, we employ UAD to screen out the OOD adversarial samples in an effort to facilitate the correct prediction of in-distribution adversarial samples by model enhanced with SSAT (Figure \ref{fig:vis}). Our method is tailor-designed for classifying medical imaging data sets with a limited number of labels and can robustly defend against various unseen attacks. 

\section{Proposed Model}

The medical image classification problem is to train a prediction function $f_{\theta}(\cdot)$ by minimizing the loss in mapping an clean image $\boldsymbol{x}\in \mathcal{X}$ to its true label $\boldsymbol{y}\in \mathcal{Y}$. Due to the existence of adversarial samples $\boldsymbol{x}'\in\mathcal{X}'$, it is necessary to introduce a detection function $g_{\phi}(\cdot)$ that can distinguish whether an input of $f_\theta$ is perturbed by an adversary. Ideally $g_\phi$ takes inputs from both $\mathcal{X}$ and $\mathcal{X}'$, rejects all $\boldsymbol{x}'$ from $\mathcal{X}'$, then $f_\theta$ only takes $\boldsymbol{x}$ from $\mathcal{X}$ to make predictions. A promising solution is to design a UAD function $g_\phi$ to reject all OOD adversarial samples from $\mathcal{X}'$. However, it is a challenging task since some of them (i.e., in-distribution adversarial samples) are very close to clean images (Figure \ref{fig:vis}). As such, a supervised prediction function $f_\theta$ that is capable of correctly classifying those adversarial samples using limited labeled training set is also indispensable for maximizing the defense effectiveness.

\begin{figure}[t]
\centering
\includegraphics[width=0.8\textwidth]{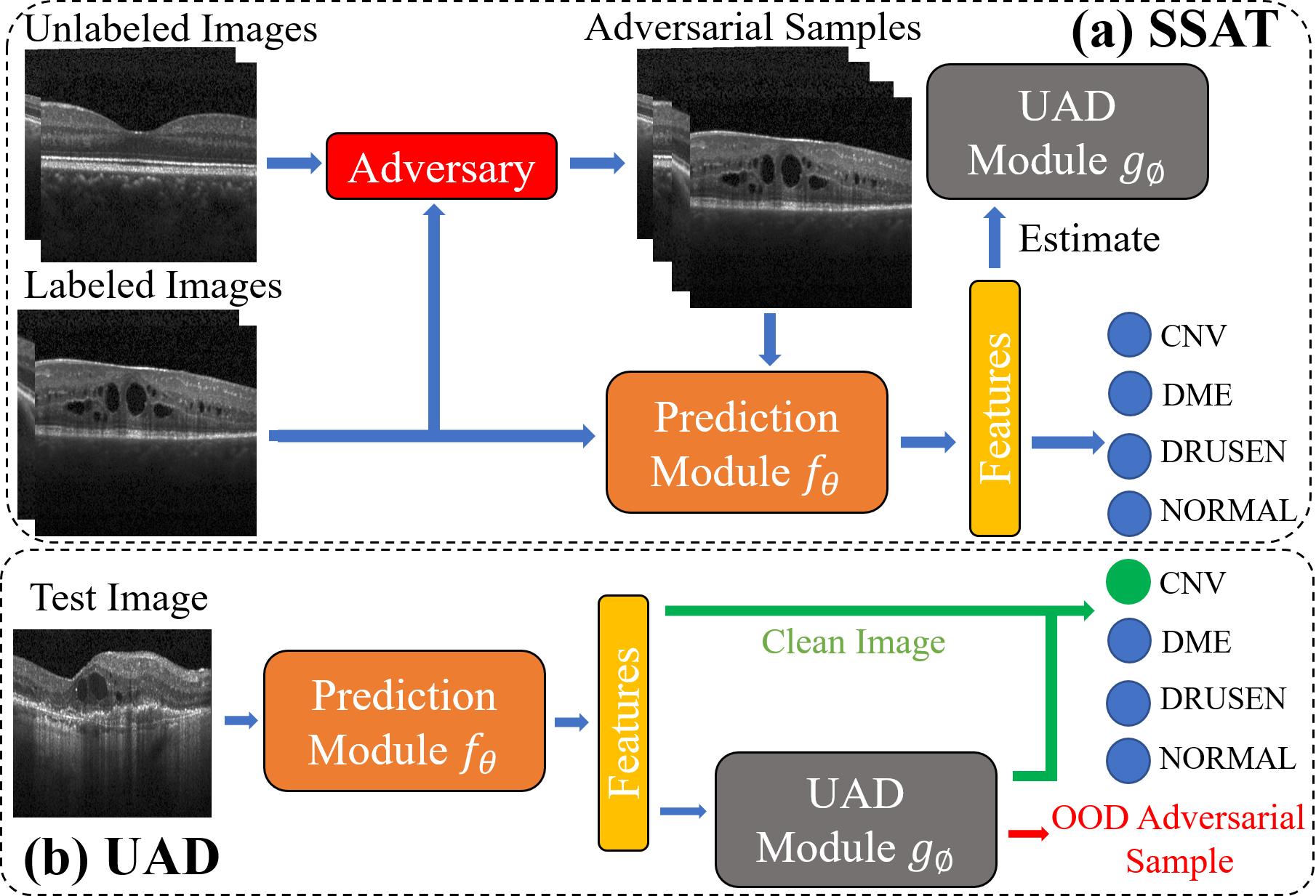}
\caption{The proposed robust OCT imaging classification system equipped with SSAT and UAD modules.}
\label{fig:framework}
\end{figure}

Figure \ref{fig:framework} illustrates our adversarial defense approach. During training (Figure \ref{fig:framework}a), we learn the robust feature representation via SSAT for both prediction and UAD modules. During inference (Figure \ref{fig:framework}b), given an unseen test image (clean or adversarial), the system extracts the feature as the input for UAD module. The test image is rejected if it is detected as a OOD adversarial sample, otherwise it continues to the loss layer to predict a class label. We describe the technical detail of SSAT and UAD modules in the following subsections.





\subsection{Semi-supervised Adversarial Training}

Adversarial training (AT) \cite{goodfellow2014explaining} is a powerful way to improve the adversarial robustness of a prediction module when the labeled training set is abundant. Recently adversarial samples generated from unlabeled data with pseudo labels have been shown to be valuable for improving the adversarial robustness \cite{stanforth2019labels}. In training the prediction module $f_\theta$ with labeled images, we use the supervised AT, i.e.,
 $   \mathcal{L}_{\text {sup}}(\theta)=\underset{x \in \mathcal{X}}{\mathbb{E}} \sup _{x^{\prime} \in \mathcal{N}_{\epsilon}(x)} \operatorname{xent}\left(y, f_{\theta}\left(x^{\prime}\right)\right),$
where $\mathcal{N}_{\epsilon}(x)$ denotes the neighborhood of a clean image $x$ and $||x-x'||_{\infty} < \epsilon$. The inner maximization can be approximated by any available attack method, such as PGD and FGSM. For training with unlabeled images, we first find their pseudo labels $\hat{y}(x)$ predicted by $f_\theta$, followed by AT, i.e., minimizing
 $   \mathcal{L}_{\text {unsup }}(\theta)=\underset{x \in \mathcal{X}'}{\mathbb{E}}  \sup _{x^{\prime} \in \mathcal{N}_{\epsilon}(x)} \operatorname{xent}\left(\hat{y}(x), f_{\theta}\left(x^{\prime}\right)\right).$
We then minimize the loss function in Eq. \ref{eq:ssat} to perform SSAT in an effort to enhance model's adversarial robustness:
\begin{align}
    \label{eq:ssat}
    \mathcal{L_{\text {semi-sup }}}(\theta) = \mathcal{L}_{\text {sup }}(\theta) + \lambda \mathcal{L}_{\text {unsup }}(\theta),
\end{align}
where $\lambda$ is a hyper-parameter tunned according to the relative abundances of labeled and unlabeled training data \cite{stanforth2019labels}.

\subsection{Unsupervised Adversarial Detection}

To filter out OOD adversarial samples $\boldsymbol{x}'$ from being fed into $f_\theta$, we design an UAD module $g_\phi$ with the goal to exclude the majority of unseen adversarial samples $\boldsymbol{x}'\in\mathcal{X}'$, and simultaneously prevent $\boldsymbol{x}\in \mathcal{X}$ from being erroneously rejected. As shown in Figure \ref{fig:vis}, the clean images have a different distribution from adversarial samples classified into the same class (color). Inspired by this observation, we estimate a probability density only for clean images as the UAD module and reject images deviating away from this density as OOD adversarial samples. Unlike the detection methods described in \cite{ma2018characterizing,ma2019understanding,feinman2017detecting} our proposed UAD is completely unsupervised that does not need to estimate the adversarial density in whatever way thus is not limited to detecting the adversarial samples from the known attack types. Specifically, let $\boldsymbol{Z}$ be the latent feature extracted from the penultimate layer of $f_\theta$ using $\boldsymbol{x}$ as input and we employ a Gaussian mixture model (GMM) for UAD module $g_\phi$. Let $\boldsymbol{\mu}_{ij}\in \mathbb{R}^n$ and $\boldsymbol{\Sigma}_{ij} \in \mathbb{R}^{n\times n}$ represents the mean and covariance matrix of the $j$th Gaussian component of class $i$, respectively. For a single class, given all features extracted from clean training samples $\boldsymbol{Z} =\{\boldsymbol{z}_1,\ldots, \boldsymbol{z}_n \}$, we can estimate parameters of the GMM using EM algorithm. The high dimension of $\boldsymbol{Z}$ may cause numerical issues during training thus a small non-negative regularization is added to the diagonal of the covariance matrices to alleviate these issues \cite{scikit-learn}. 



\subsection{Adversarial Risk Evaluation}\label{sec:eva}
We propose a new adversarial risk evaluation measure for comparing systems performance in terms of adversarial defense. We assess the risk derived from clean images based on the following intuition: 1) a clean image incurs no risk if it can be correctly classified, 2) a clean image being rejected by the UAD incurs risk $r_{cln}^{uad}$, 3) a clean image being accepted by the UAD but misclassified by prediction model incurs risk $r_{cln}^{prd}$. Assume that for clean images, the number of accepted images that incorrectly predicted is $N_{cln}^{inc}(f,g)$, and the number of clean images being rejected is $N_{cln}^{rej}(g)$, the risk derived from misclassifying (first term) and erroneously rejecting (second term) clean images is calculated as
$R_{cln}(f, g) = N_{cln}^{inc}(f,g) \cdot r_{cln}^{prd} + N_{cln}^{rej}(g)\cdot r_{cln}^{uad}.$
If only $f$ is used to make predictions (without UAD), the second term is zeroed out. Lets denote $N_{cln}^{inc}(f)$ as the number of clean images being misclassified by $f$, then the risk is calculated as
$    R_{cln}(f) = N_{cln}^{inc}(f) \cdot r_{cln}^{prd}.$
 
\begin{figure}[t]
\centering
\includegraphics[width=\textwidth]{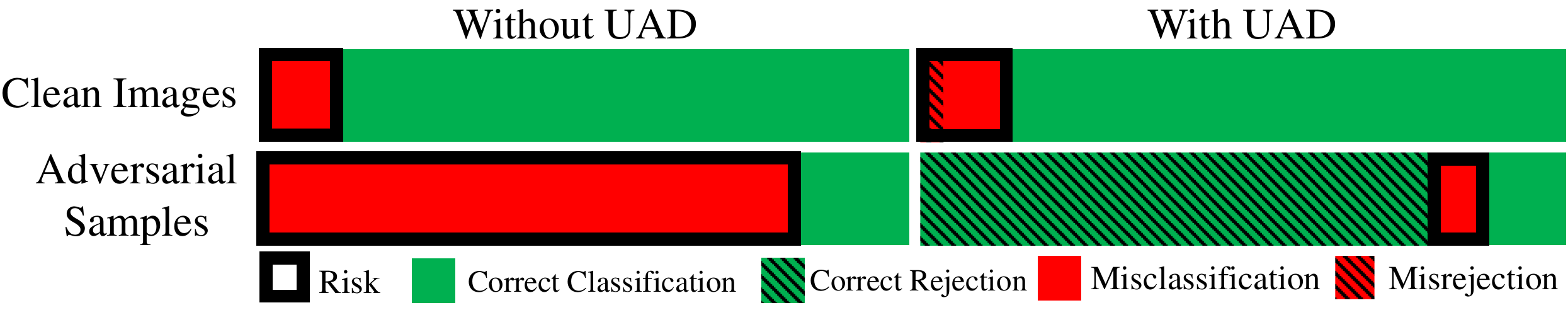}
\caption{An illustration of assessing systems adversarial risk. Note the system with UAD on the right exhibits a much lower risk represented by smaller red zones.}
\label{fig:eval}
\end{figure}


Similarly for adversarial samples, we have the following intuition: 1) being correctly rejected by UAD or bypassed but correctly classified incurs no risk, 2) being erroneously accepted by UAD but misclassified incurs a risk $r_{adv}^{prd}$. Assume the number adversarial samples in 2) is $N_{adv}^{inc}(f,g)$, the risk derived from adversarial samples is calculated as
 $   R_{adv}(f, g) =  N_{adv}^{inc}(f,g)\cdot r_{adv}^{prd}.$
When only $f$ is used to make predictions (without UAD), since $N_{adv}^{inc}(f,g) =N_{adv}^{inc}(f)$ and $N_{adv}^{inc}(f)$ is the number of misclassified adversarial samples, the risk is calculated as
 $   R_{adv}(f) =  N_{adv}^{inc}(f)\cdot r_{adv}^{prd}.$
The total risk, incurred by both clean and adversarial samples, thus can be calculated by $R = R_{cln} + R_{adv}$. The value of different risks ($r_{cln}^{uad}, r_{cln}^{prd},r_{adv}^{prd}.$) are determined empirically, then we have the risk measures for AI systems with UAD $R(f,g)$ and without UAD $R(f)$ as follows:  
\begin{align}
    \label{eq:risk}
    R(f,g) &= N_{cln}^{inc}(f,g) + N_{cln}^{rej}(g) + N_{adv}^{inc}(f,g)\\
    \label{eq:risk2}
    R(f) &= N_{cln}^{inc}(f) +  N_{adv}^{inc}(f).
\end{align}

This evaluation measures are illustrated in Figure~\ref{fig:eval}. Using the above equations, we can assess and compare average adversarial risks $r = R(f,g)/N$ between UAD based and not UAD based adversarial defense approaches.


\section{Experiments and Results}
We use experiments to demonstrate that: 1): The SSAT module can significantly increase model's adversarial robustness without compromising classification performance of clean images. 2) The UAD module can detect and exclude a majority of successful adversarial examples. 3) Our medical imaging AI system (UAD + SSAT) minimizes adversarial risk compared to other existing AI systems.
\\\\
\noindent\textbf{Dataset and Experiment Settings} The experiments are conducted on a public retinal OCT image dataset, originally released in \cite{kermany2018identifying}. It contains 84,495 images taken from 4,686 patients with 4 classes: choroidal neovascularization (CNV), diabetic macular edema (DME), drusen, and normal. To demonstrate the advantages of using unlabeled images for semi-supervised training (\ref{eq:ssat}), we randomly sample 4,000 images for training, 1,000 images for testing and additional 1,000 images as the unlabeled dataset for SSAT. The 4 classes are balanced in each data set. Following the standard prepossessing procedure \cite{goodfellow2014explaining}, all images are center-cropped to $224 \times 224$ and all pixels are scaled to [0,1]. For AT and SSAT, we augment the data set by generating adversarial samples for each mini-batch using FGSM with a uniformly sampling perturbation from the interval [0.001,0.003]. The number of adversarial and clean images remains $1:1$ within each mini-batch. We use ResNet-18 \cite{he2016deep} pre-trained with ImageNet to learn robust feature representations against adversarial attacks. The networks are trained with the SGD optimizer for 10 epochs with a batch size of 64. We set $\lambda = 5$ for SSAT Eq. \ref{eq:ssat} as in \cite{stanforth2019labels}.

\noindent\textbf{SSAT Performance} We evaluate class prediction performance under the most challenging threat: 'white-box' setting \cite{carlini2017towards}. Compared to the benign 'black-box' setting, the adversary possesses complete knowledge of the target model, including architecture and model parameters. We compare our SSAT with three baseline methods in terms of classification accuracy: natural training (NT) with cross-entropy loss, adversarial training (AT) with cross-entropy loss \cite{goodfellow2014explaining} and natural training with guided complement entropy (GCE) loss \cite{chen2019improving}. The 1,000 attacks are crafted by 1-step FGSM, 10-step PGD, and C\&W based on the test set.

\begin{figure}[ht]
\centering
\includegraphics[width=\textwidth]{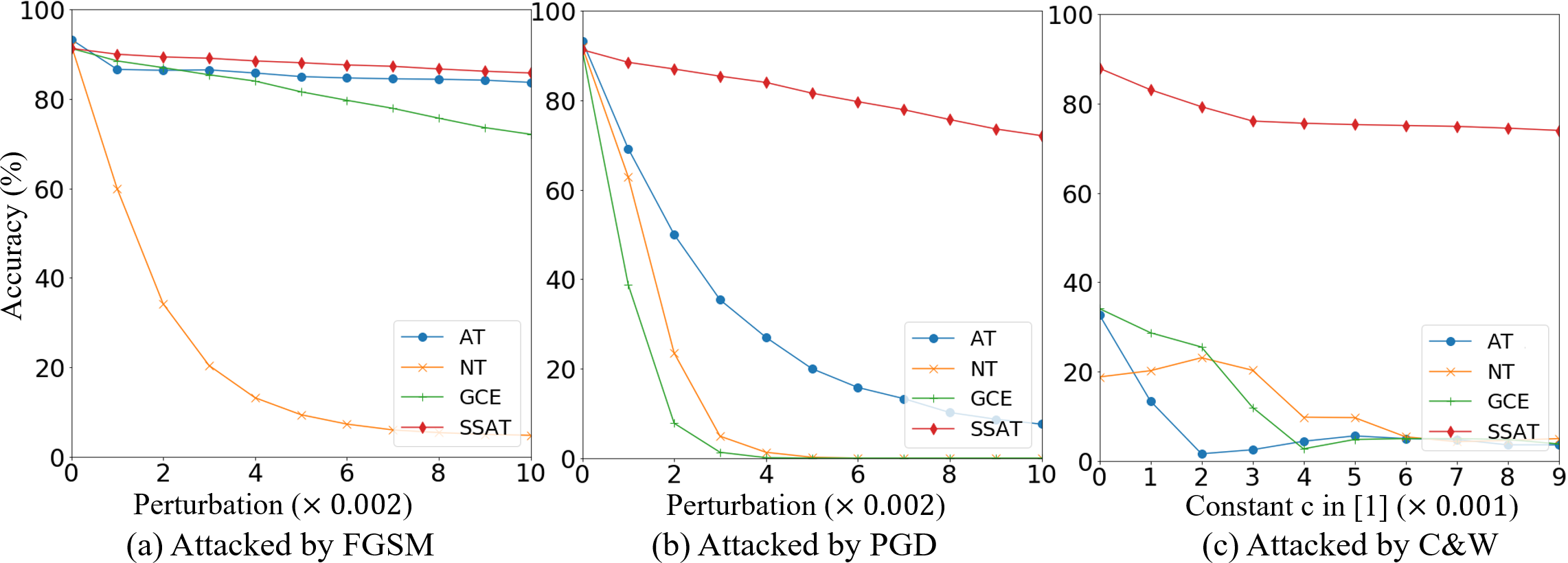}
\caption{The supervised prediction accuracy of the four trained models on 1000 adversarial examples crafted by FGSM, PGD, C\&W with an increasing perturbation budget and constant c.}
\label{fig:result_c}
\end{figure}

Figure \ref{fig:result_c} demonstrates that SSAT markedly outperforms other baselines in all white-box attack settings while maintaining a comparable or better performance on the clean image classification (when the perturbation budget is zero). The NT appears very susceptible to easy attacks generated using FGSM with a very small perturbation budget whereas GCE and AT demonstrate a solid performance against easy attacks but fail under strong attacks such as those generated using PGD and C\&W. For AT, label scarcity has significantly limited its adversarial generalizability. For GCE, widening the gap in the manifold between different classes may not work well for medical images due to significant overlaps in both the fore- and backgrounds. 

\begin{table}[ht]
	\begin{center}
		
		\begin{tabular}{C{1.5cm}|C{1.5cm}C{1.5cm}C{1.5cm}C{1.5cm}C{1.5cm}|C{1.5cm}}
			\hline
			Classes & CNV            & DME            & DRUSE          & NORMAL         & Average        & \# cases \\ \hline
			NT      & 0.897          & 0.802          & 0.852          & 0.859          & 0.852          & 885      \\ 
			GCE     & 0.943          & 0.902          & 0.930          & 0.931          & 0.927          & 970      \\ 
			AT      & 0.890          & 0.932          & 0.841          & 0.903          & 0.892          & 580      \\ 
			SSAT    & \textbf{0.965} & \textbf{0.987} & \textbf{0.967} & \textbf{0.974} & \textbf{0.973} & 136      \\ \hline
		\end{tabular}
	\end{center}
	\caption{Performance comparison using AUPRC under PGD attack with a perturbation $\epsilon=0.005$. The last column shows the number of successful adversarial samples.}
	\label{table:detection}
\end{table}

\begin{table}[th]
	\begin{center}
		\begin{tabular}{C{4.0cm}|C{1.2cm}C{1.2cm}C{1.2cm}C{1.2cm}C{1.2cm}}
			\hline
			Method            & NT             & GCE            & AT            & SSAT           & SSAT*          \\ \hline
			Adversarial Risk w/o UAD & 0.965          & 1.057          & 0.647          & 0.223          & 0.912          \\ 
			Adversarial Risk w. UAD     & \textbf{0.892} & \textbf{0.713} & \textbf{0.634} & \textbf{0.215} & \textbf{0.450} \\ \hline
			Prediction Accuracy    & 11.5\%          & 0.3\%           & 42\%            & 86.4\%          & 17.5\%          \\ \hline
		\end{tabular}
	\end{center}
	\caption{Systems risk under PGD attack with perturbation $\epsilon= 0.005$. SSTA* indicates the risk under a stronger PGD attack $\epsilon= 0.01$. }
	\label{table:risk}
	
\end{table}

\noindent\textbf{UAD Performance} We use features extracted from 4000 clean images in the training set to estimate mixture model density for UAD. Then the 1000 images from test set and its successful adversarial counterparts are used for assessing performance of UAD. As shown in Table \ref{table:detection}, UAD is more effective in detecting and excluding adversarial samples evident by the higher AUPRC values among all classes. Furthermore SSAT is more effective than other training strategies, i.e., NT, AT or GCE. Since the classes of clean images and successful attacks are highly imbalanced (136:1000), AUPRC is a suitable metric for performance evaluation \cite{saito2015precision}. The average AUPRC value of 0.973 shows the proposed UAD can correctly filter out a vast majority of OOD adversarial samples.

\noindent\textbf{Comparison of Adversarial Risks} Finally, we demonstrate that UAD complementing with SSAT give rise to the lowest adversarial risk in terms of the new measure proposed in Eq. \ref{eq:risk}\&\ref{eq:risk2}. In Table \ref{table:risk}, it is clear that UAD based systems have consistently lower risks compared to those are not, regardless of the training methods used. Note that the reduction of risk is not significant for SSAT against PGD attacks with a smaller budget ($\epsilon$ = 0.005). The main reason is that these adversarial samples are relatively weak (highest class prediction accuracy of 86.4\% in the last row) that SSAT can successful predict their labels without the need for UAD. After we double the perturbation budget of PGD attack ($\epsilon$ = 0.005), as shown in the last column, the adversarial risk decreases by half (from 0.912 to 0.450) with UAD, highlighting the striking robustness of our system against stronger PGD attacks comparing with those without UAD.

\section{Conclusions}
We propose to enhance the robustness of medical image AI system via UAD complemented with SSAT. The former is to imbue the system with robustness against unseen OOD adversarial samples whereas the latter mitigate the label scarcity problem in training a robust classifier for predicting in-distribution adversarial samples. Though experiments our system demonstrates a superior performance in adversarial defense to competing techniques.

%
%
\bibliographystyle{splncs04}
\bibliography{mybibliography}

\end{document}